\documentclass[lettersize,journal]{IEEEtran}
\usepackage{amsmath,amsfonts}
\usepackage{algorithmic}
\usepackage{algorithm}
\usepackage{array}
\usepackage[caption=false,font=normalsize,labelfont=sf,textfont=sf]{subfig}
\usepackage{textcomp}
\usepackage{stfloats}
\usepackage{url}
\usepackage{verbatim}
\usepackage{graphicx}
\usepackage{cite}

\usepackage{color}
\usepackage{multirow}
\usepackage{amsmath}
\usepackage{amssymb}
\DeclareRobustCommand\red{\textcolor{red}}
\DeclareRobustCommand\blue{\textcolor{blue}}
\usepackage{times}
\usepackage{epsfig}

\begin{document}

\title{Visformer: The Vision-friendly Transformer}


\author{Zhengsu Chen,  Lingxi Xie, Jianwei Niu, \textit{Senior Member}, \textit{IEEE}, \\ Xuefeng Liu, Longhui Wei, Qi Tian, \textit{Fellow}, \textit{IEEE} 
\thanks{Zhengsu Chen and Xuefeng Liu are with the School of Computer Science and Engineering, Beihang University, {\rm 100191}, Beijing, China (e-mail: danczs@buaa.edu.cn; liu\_xuefeng@buaa.edu.cn)}
\thanks{Lingxi Xie is with Johns Hopkins University, Baltimore, USA (e-mail: 198808xc@gmail.com)}
\thanks{Jianwei Niu is with Hangzhou Innovation Institute of Beihang University, Zhengzhou University, and Beihang University, {\rm 100191}, Beijing,  China (e-mail: @niujianwei@buaa.edu.cn)}
\thanks{Longhui wei is with the University of Science and Technology of China, {\rm 230026}, Hefei, China (e-mail:longhuiwei@pku.edu.cn)}
\thanks{Qi Tian is with Xidian University, {\rm 710126}, Xi'an, China (e-mail: wywqtian@gmail.com)}

}

\markboth{}%
{Chen \MakeLowercase{\textit{et al.}}: VisformerV2: The more Vision-friendly Transformer}

\IEEEpubid{0000--0000/00\$00.00~\copyright~2021 IEEE}

\maketitle

\begin{abstract}
The past few years have witnessed the rapid development of applying the Transformer module to vision problems. While some researchers have demonstrated that Transformer-based models enjoy a favorable ability of fitting data, there are still growing number of evidences showing that these models suffer over-fitting especially when the training data is limited. This paper offers an empirical study by performing step-by-step operations to gradually transit a Transformer-based model to a convolution-based model. The results we obtain during the transition process deliver useful messages for improving visual recognition. Based on these observations, we propose a new architecture named Visformer, which is abbreviated from the `Vision-friendly Transformer'. With the same computational complexity, Visformer outperforms both the Transformer-based and convolution-based models in terms of ImageNet classification and object detection performance, and the advantage becomes more significant when the model complexity is lower or the training set is smaller. The code is available at \url{https://github.com/danczs/Visformer}.
\end{abstract}


\section{Introduction}
\IEEEPARstart{I}{n} the past decade, convolution used to play a central role in the deep learning models~\cite{lecun2015deep, simonyan2014very, szegedy2015going, he2016deep} for visual recognition. This situation starts to change when the Transformer~\cite{vaswani2017attention}, a module that originates from natural language processing~\cite{vaswani2017attention, devlin2018bert,radford2018improving}, is transplanted to the vision scenarios. It was shown in the ViT model~\cite{dosovitskiy2020image} that an image can be partitioned into a grid of patches and the Transformer is directly applied upon the grid as if each patch is a visual word. ViT requires a large amount of training data (\textit{e.g.}, the ImageNet-21K~\cite{deng2009imagenet} or the JFT-300M dataset), arguably because the Transformer is equipped with long-range attention and interaction, and is prone to over-fitting. The follow-up efforts~\cite{touvron2020training} improved ViT to some extent, but these models still perform badly especially under limited training data or moderate data augmentation compared with convolution-based models.

On the other hand, vision Transformers can achieve much better performance than convolution-based models when trained with large amount of data. Namely, vision Transformers have higher `upper-bound' while convolution-based models are better in `lower-bound'. Both upper-bound and lower-bound are important properties for neural networks. Upper-bound is the potential to achieve higher performance and lower-bound enables networks to perform better when trained with limited data or scaled to different complexity.

\begin{table}
\begin{center}
\setlength{\tabcolsep}{0.12cm}
\caption{The comparison among ResNet-50, DeiT-S, and the proposed Visformer-S model on ImageNet. }
\begin{tabular}{|l|l|c|c|c|}
\hline
\multicolumn{2}{|l|}{Network} & ResNet-50 & DeiT-S & Visformer-S \\
\hline\hline
\multicolumn{2}{|l|}{FLOPs (G)} & 4.1 & 4.6 & 4.9 \\
\hline
\multicolumn{2}{|l|}{Parameters (M)} & 25.6 & 21.8 & 40.2 \\
\hline\hline
Full & base setting  & 77.43 & 63.12 & 77.20 \\
\cline{2-5}
data & elite setting & 78.73 & 80.07 & 82.19 \\
\hline
Part of & 10\% labels & 58.37 &40.41 & 58.74 \\
\cline{2-5}
data & 10\% classes  & 89.90  &80.06  &  90.06 \\
\hline
\end{tabular}
\end{center}
\label{tab:introduction}
\end{table}

Based on the observation of lower-bound and upper-bound on Transformer-based and convolution-based networks, the main goal of this paper is to identify the reasons behind the difference, by which we can design networks with higher lower-bound and upper-bound. The gap between Transformer-based and convolution-based networks can be revealed with two different training settings on ImageNet. The first one is the base setting. It is the standard setting for convolution-based models, \textit{i.e.}, the training schedule is shorter and the data augmentation only contains basic operators such as random-size cropping~\cite{szegedy2016rethinking} and flipping. The performance under this setting is called \textbf{base performance} in this paper. The other one is the training setting used in~\cite{touvron2020training}. It is carefully tuned for Transformer-based models, \textit{i.e.}, the training schedule is longer and the data augmentation is stronger (\textit{e.g.}, RandAugment~\cite{cubuk2020randaugment}, CutMix~\cite{yun2019cutmix}, \textit{etc.}, have been added). We use the \textbf{elite performance} to refer to the accuracy produced by it.

We take DeiT-S~\cite{touvron2020training} and ResNet-50~\cite{he2016deep} as the examples of Transformer-based and convolution-based models. As shown in Table~\ref{tab:introduction}, Deit-S and ResNet-50 employ comparable FLOPs and parameters. However, they behave very differently trained on the full data under these two settings. Deit-S has higher elite performance, but changing the setting from elite to base can cause a $10\%+$ accuracy drop for DeiT-S. ResNet-50 performs much better under the base setting, yet the improvement for the elite setting is merely $1.3\%$. This motivates us to study the difference between these models. 
\IEEEpubidadjcol 
With these two settings, we can roughly estimate the lower-bound and upper-bound of the models. The methodology we use is to perform step-by-step operations to gradually transit one model into another, by which we can identify the properties of modules and designs in these two networks. The entire transition process, taking a total of $8$ steps, is illustrated in Figure~\ref{fig:transition}.

Specifically, from DeiT-S to ResNet-50, one should (i) use global average pooling (not the classification token), (ii) introduce step-wise patch embeddings (not large patch flattening), (iii) adopt the stage-wise backbone design, (iv) use batch normalization~\cite{ioffe2015batch} (not layer normalization~\cite{ba2016layer}), (v) leverage $3\times3$ convolutions, (vi) discard the position embedding scheme, (vii) replace self-attention with convolution, and finally (viii) adjust the network shape (\textit{e.g.}, depth, width, \textit{etc.}). After a thorough analysis on the reasons behind the results, we absorb all the factors that are helpful to visual recognition and derive the \textbf{Visformer}, \textit{i.e.}, the Vision-friendly Transformer.

Evaluated on ImageNet classification, Visformer claims better performance than the competitors, DeiT and ResNet, as shown in Table~\ref{tab:introduction}. With the elite setting, the Visformer-S model outperforms DeiT-S and ResNet-50 by $2.12\%$ and $3.46\%$, respectively, under a comparable model complexity. Different from Deit-S, Visformer-S also survives two extra challenges, namely, when the model is trained with 10\% labels (images) and 10\% classes. Visformer-S even performs better than ResNet-50, which reveals the high lower-bound of Visformer-S. Additionally, for tiny models, Visformer-Ti significantly outperforms Deit-Ti by more than 6\%.

The contribution of this paper is three-fold. \textbf{First}, for the first time, we introduce the lower-bound and upper-bound to investigate the performance of Transformer-based vision models. \textbf{Second}, we close the gap between the Transformer-based and convolution-based models by a gradual transition process and thus identify the properties of the designs in the Transformer-based and convolution-based models. \textbf{Third}, we propose the Visformer as the final model that achieves satisfying lower-bound and upper-bound.

The preliminary version of this paper appeared as~\cite{chen2021visformer}. In the extended version, we further explore the recently proposed work and provide more experiments and analysis. The  main improvements over the preliminary version are summarized as follows:
\begin{itemize}
    \item We optimize the architecture of Visformer according to the experimental observations and propose VisformerV2 which substantially outperforms the old version.
    \item We analyze the overflow problem when utilizing half-precision in Transformers and propose an efficient method to avoid overflow without degrading the performance.
    \item We generalize Visformer to downstream vision tasks and observe consistent improvements.
\end{itemize}
\section{Related work}
Image classification is a fundamental task in computer vision. In the deep learning era, the most popular method is to use deep neural networks~\cite{krizhevsky2012imagenet,simonyan2014very,he2016deep}. One of the fundamental units to build such networks is convolution, where a number of convolutional kernels are used to capture repeatable local patterns in the input image and intermediate data. To reduce the computational costs as well as alleviate the risk of over-fitting, it was believed that the convolutional kernels should be of a small size, \textit{e.g.}, $3\times3$. However, this brings the difficulty for faraway contexts in the image to communicate with each other -- this is partly the reason that the number of layers has been increasing. Despite stacking more and more layers, researchers consider another path which is to use attention-based approaches to ease the propagation of visual information.

Since Transformers achieved remarkable success in natural language processing (NLP)~\cite{vaswani2017attention,devlin2018bert,radford2018improving}, many efforts have been made to introduce Transformers to vision tasks. These works mainly fall into two categories. The first category consists of pure attention models~\cite{ramachandran2019stand,hu2019local, zhao2020exploring, chen2020generative, dosovitskiy2020image, touvron2020training, wang2021pyramid}. These models usually only utilize self-attention and attempt to build vision models without convolutions. However, it is computationally expensive to relate all pixels with self-attention for realistic full-sized images. Thus, there has some interest in forcing self-attention to only concentrate on the pixels in local neighborhoods (\textit{e.g.}, SASA~\cite{ramachandran2019stand}, LRNet~\cite{hu2019local}, SANet~\cite{zhao2020exploring}). These methods replace convolutions with local self-attentions to learn local relations and achieve promising results. However, it requires complex engineering to efficiently apply self-attention to every local region in an image. Another way to solve the complexity problem is to apply self-attention to reduced resolution. These methods either reduce the resolution and color space first~\cite{chen2020generative} or regard image patches rather pixels as tokens (\textit{i.e.}, words)~\cite{dosovitskiy2020image, touvron2020training}. However, resolution reduction and patch flattening usually make it more difficult to utilize the local prior in natural images. Thus, these methods usually obtain suboptimal results~\cite{chen2020generative} or require huge dataset~\cite{dosovitskiy2020image} and heavy augmentation~\cite{touvron2020training}.  

The second category contains the networks built with not only self-attentions but also convolutions. Self-attention was first introduced to CNNs by non-local neural networks~\cite{wang2018non}. These networks aim to capture global dependencies in images and videos. Note that non-local neural networks are inspired by the classical non-local method in vision tasks~\cite{buades2005non} and unlike those in Transformers, the self-attentions in non-local networks are usually not equipped with multi-heads and position embedding~\cite{wang2018non, cao2019gcnet, li2020neural}. Afterwards, Transformers achieve remarkable success in NLP tasks~\cite{devlin2018bert,radford2018improving} and, therefore, self-attentions that inherits NLP settings (\textit{e.g.}, multi-heads, position encodings, classification token, \textit{etc.}) are combined with convolutions to improve vision tasks~\cite{ramachandran2019stand, bello2019attention, bello2021lambdanetworks}. A common combination is to utilize convolutions first and apply self-attention afterwards~\cite{dosovitskiy2020image, srinivas2021bottleneck}. \cite{dosovitskiy2020image} builds hybrids of self-attention and convolution by adding a ResNet backbone before Transformers. Afterwards, more and more methods are proposed to combine self-attention and convolution. Besides utilizing convolution in early layers~\cite{vaswani2021scaling}, BotNet~\cite{srinivas2021bottleneck} designs bottleneck cells for self-attention. Conformer~\cite{peng2021conformer} fuses the feature of a convolution neural network and a Transformer with the feature coupling unit to combine the global and local representations. CvT~\cite{wu2021cvt} introduces convolution to the feature projection of self-attention, by which the query, key and value in the self-attention can capture the local information. CoAtNet~\cite{dai2021coatnet} unifies depthwise convolution with self-attention and vertically stacks convolution layers and self-attention layers to improve the generalization, capacity and efficiency. However, these methods usually combine convolution and self-attention empirically or heuristically. Our method, in contrast, explores the full process of converting a Transformer to a convolution neural network.

There is also work that studies scaling or training vision Transformer. CaiT~\cite{touvron2021going} find that it is very efficient to scale up vision Transformer in depth dimension with LayerScale. AutoFormer~\cite{chen2021autoformer} builds a super Transformer network by which they can evaluate the different designs and search efficient Transformer architecture. Zhai \textit{et al.}~\cite{zhai2021scaling} observe that most vision Transformers can benefit from increasing compute resources and larger dataset. They suggest scaling up compute, data and model together. Steiner \textit{et al.}~\cite{steiner2021train} further study data, augmentation and regularization in vision Transformer and finds that augmentation can yield the same performance as that trained on an order of magnitude more data. 

Additionally, self-attention has been used in many downstream vision tasks (detection~\cite{carion2020end}, segmentation~\cite{chen2021transunet}) and low vision tasks~\cite{chen2020pre}. These tasks usually utilize much larger input resolution than classification. For example, the frameworks in COCO~\cite{lin2014microsoft} usually utilize $1280 \times 800$ inputs. This is a critical problem for vision Transformer, since the complexity increases quadratically with pixel numbers. The widely used solution is adopting sliding windows to capture local patterns and building extra pipelines for information exchange among the windows. Swin Transformer~\cite{liu2021swin} shifts the windows alternately in different layers, by which the tokens can build long-distance relations as the depth increases. CSWin Transformer~\cite{dong2021cswin} further develops the cross-shaped self-attention mechanism to ensure that the windows can access the global feature in one dimension. MSG-Transformer~\cite{fang2021msg}, by contrast, exchanges the local information by messenger tokens.

\section{Methodology}
\label{methodology}

\subsection{Transformer-based and convolution-based visual recognition models}
\label{methodology:baselines}

Recognition is the fundamental task in computer vision. This work mainly considers image classification, where the input image is propagated through a deep network to derive the output class label. Most deep networks are designed in a hierarchical manner and composed of a series of layers.



We consider two popular layers named convolution and Transformer. Convolution originates from the intuition to capture local patterns which are believed more repeatable than global patterns. It uses a number of learnable kernels to compute the responses of the input to different patterns, for which a sliding window is moved along both axes of the input data and the inner-product between the data and kernel is calculated. In this paper, we constrain our study in the scope of residual blocks, a combination of $2$ or $3$ convolutional layers and a skip-connection. Non-linearities such as activation and normalization are inserted between the neighboring convolutional layers. 

On the other hand, Transformer originates from natural language processing and aims to frequently formulate the relationship between \textit{any} two elements (called tokens) even when they are far from each other. This is achieved by generating three features for each token, named the query, key, and value, respectively. Then, the response of each token is calculated as a weighted sum over all the values, where the weights are determined by the similarity between its query and the corresponding keys. This is often referred to as multi-head self-attention (MHSA), followed by other operations including normalization and linear mapping.

Throughout the remaining part, we consider DeiT-S~\cite{touvron2020training} and ResNet-50~\cite{he2016deep} as the representative of Transformer-based and convolution-based models, respectively. Besides the basic building block, there are also differences in design, \textit{e.g.}, ResNet-50 has a few down-sampling layers that partition the model into stages, but the number of tokens remains unchanged throughout DeiT-S. The impact of these details will be elaborated in Section~\ref{methodology:transition}.

\subsection{Settings: The base and elite performance}
\label{methodology:settings}

Although DeiT-S reports a $80.1\%$ accuracy which is higher than $78.7\%$ of ResNet-50, we notice that DeiT-S has changed the training strategy significantly, \textit{e.g.}, the number of epochs is enlarged by more than $3\times$ and the data augmentation becomes much stronger. Interestingly, DeiT-S seems to heavily rely on the carefully-tuned training strategy, and other Transformer-based models including ViT~\cite{dosovitskiy2020image} and PIT~\cite{chen2020pre} also reported their dependency on other factors, \textit{e.g.}, a large-scale training set. In what follows, we provide a comprehensive study on this phenomenon.

We evaluate all classification models on the ImageNet dataset~\cite{russakovsky2015imagenet} which has $1\mathrm{K}$ classes, $1.28\mathrm{M}$ training images and $50\mathrm{K}$ testing images. Each class has roughly the same number of training images. This is one of the most popular datasets for visual recognition.

There are two settings to optimize each recognition model. The first one is named the \textbf{base setting} which is widely adopted by convolution-based networks. Specifically, the model is trained for $90$ epochs with the SGD optimizer. The learning rate starts with $0.2$ for batch size $512$ and gradually decays to $0.00001$ following the cosine annealing function. A moderate data augmentation strategy with random-size cropping~\cite{szegedy2016rethinking} and flipping is used. The second one is named the \textbf{elite setting} which has been verified effective to improve the Transformer-based models. The Adamw optimizer with an initial learning rate of $0.0005$ for batch size $512$ is used. The data augmentation and regularization strategy is made much stronger to avoid over-fitting, for which intensive operations including RandAugment~\cite{cubuk2020randaugment}, Mixup~\cite{Zhang2017mixup}, CutMix~\cite{yun2019cutmix}, Random Erasing~\cite{zhong2020random}, Repeated Augmentation~\cite{berman2019multigrain,hoffer2020augment} and Stochastic Depth~\cite{huang2016deep} are used. Correspondingly, the training lasts $300$ epochs, much longer than that of the base setting. 

Throughout the remaining part of this paper, we refer to the classification accuracy under the base and elite settings as \textbf{base performance} and \textbf{elite performance}, respectively. We expect the numbers to provide complementary views for us to understand the studied models.

\begin{figure*}[t]
\begin{center}
\includegraphics[width=\linewidth]{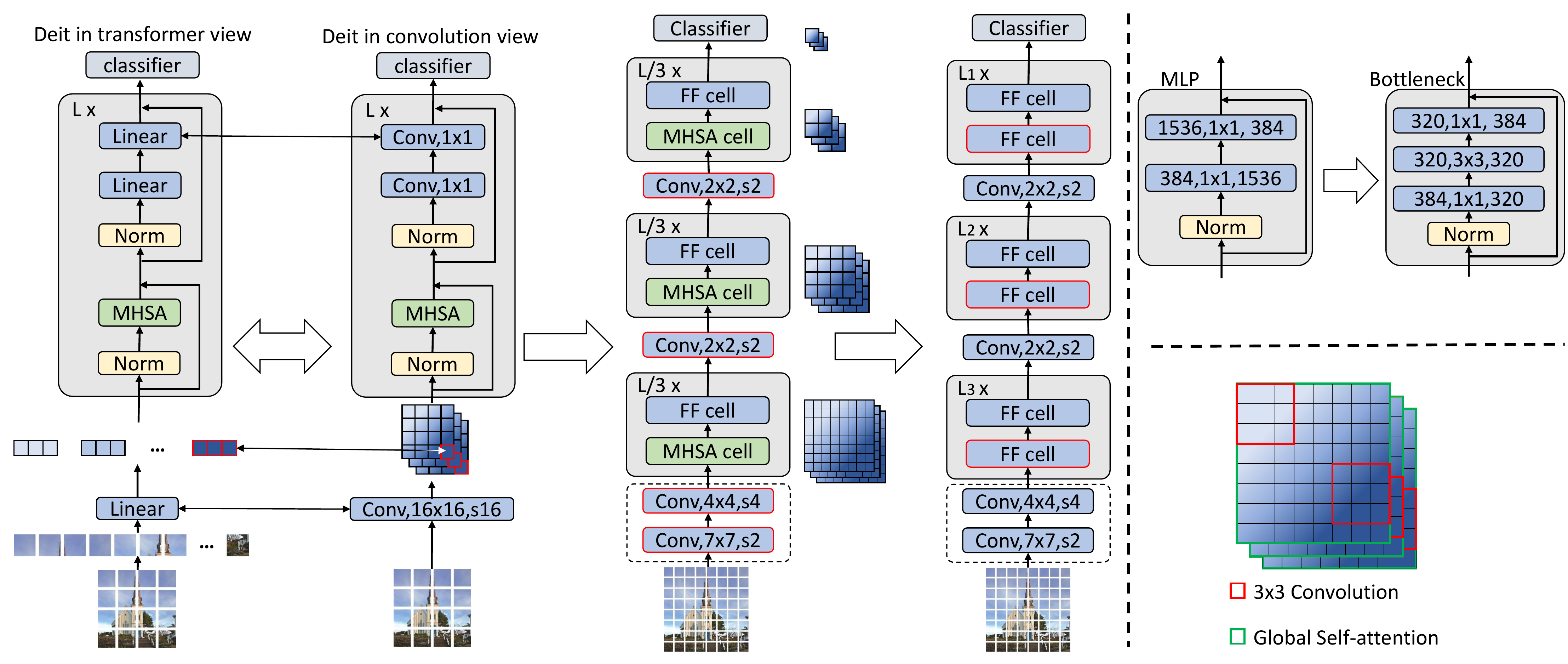}
\end{center}
\caption{The transition process that starts with DeiT and ends with ResNet-50. To save space, we only show three important movements. The first movement converts DeiT from the Transformer to convolution view (Section~\ref{methodology:transition:token}). The second movement replaces the patch flattening module with step-wise patch embedding (elaborated in Section~\ref{methodology:transition:flattening}) and introduces the stage-wise design (Section~\ref{methodology:transition:stage}) . The third movement replaces the self-attention module with convolution (Section~\ref{methodology:transition:feedforward}).
The upper-right area shows a relatively minor modifications, inserting $3\times3$ convolution (Section~\ref{methodology:transition:conv}). The lower-right area compares the receptive fields of a $3\times3$ convolution and self-attention. This figure is best viewed in color.}
\label{fig:transition}
\end{figure*}

\subsection{The transition from DeiT-S to ResNet-50}
\label{methodology:transition}

This subsection displays a step-by-step process in which we gradually transit a model from DeiT-S to ResNet-50. There are eight steps in total. The key steps are illustrated in Figure~\ref{fig:transition}, and the results, including the base and elite performance and the model statistics, are summarized in Table~\ref{tab:transition}.

\subsubsection{Using global average pooling to replace the classification token}
\label{methodology:transition:token}

The first step of the transition is to remove the classification token and add global average pooling to the Transformer-based models. Unlike the convolution-based models, Transformers usually add a classification token to the inputs and utilize the corresponding output token to perform classification, which is inherited from NLP tasks~\cite{devlin2018bert}. As a contrast, the classification features in convolution-based models are obtained by conducting global average pooling in the space dimension. 

By removing the classification token, the Transformer can be equivalent translated to the convolutional version as shown in Figure~\ref{fig:transition}. Specifically, the patch embedding operation is equivalent to a convolution whose kernel size and stride is the patch size~\cite{dosovitskiy2020image}. The shape of the intermediate features can be naturally converted from a sequence of tokens (\textit{i.e.}, words) to a bundle feature maps and the tokens become the vector in channel dimension (illustrated in Figure~\ref{fig:transition}). The linear layers in MHSA and MLP blocks are equivalent to $1\times 1$ convolutions.

The performance of the obtained network (Net1) is shown in Table~\ref{tab:transition}. As can be seen, this transition can substantially improve the base performance. Our further experiments show that adding global pooling itself can improve the base performance from 64.17\% to 69.44\%. In other words, the global average pooling operation which is widely used in convolution-based models since NIN~\cite{lin2013network}, enables the network to learn more efficiently under moderate augmentation. Furthermore, this transition can slightly improve the elite performance.

\begin{table*}
\caption{The classification accuracy on ImageNet during the transition procedure from DeiT-S to ResNet-50. Both the base setting and the elite setting are considered (for the details, see Section~\ref{methodology:settings}), and we mark the positive modifications in \red{red} and the negative modifications in \blue{blue}. Note that a modification can impact the base and elite performance differently. Though the number of parameters increases considerably at the intermediate status, the computational costs measured by FLOPs does not change significantly.}
\begin{center}
\setlength{\tabcolsep}{0.12cm}
\begin{tabular}{|l|c|c|c|c|c|c|c|}
\hline
Model Name & \textbf{added} & \textbf{removed} & \textbf{base} perf. & \textbf{elite} perf. & FLOPs (G) & Params (M) \\
\hline\hline
DeiT-S & \multicolumn{2}{c|}{--} & 64.17 & 80.07 & 4.60 & 22.1\\
\hline
Net1 & global average pooling & classification token & 69.81 (\red{+5.64}) & 80.16 (\red{+0.09}) & 4.57 & 22.0 \\
Net2 & step-wise embeddings & large patch embedding & 73.01 (\red{+3.20}) & 81.35 (\red{+1.19}) & 4.77 & 23.9 \\
Net3 & stage-wise design & -- & 75.76 (\red{+2.75}) & 80.19 (\blue{-1.14}) & 4.79 &39.5\\
Net4 & batch norm & layer norm & 76.49 (\red{+0.73}) & 80.97 (\red{+0.78}) & 4.79 & 39.5\\
Net5 & $3\times3$ convolution & -- & 77.37 (\red{+0.88}) & 80.15 (\blue{-0.82}) &4.76 &39.2\\
Net6 & -- & position embedding & 77.31 (\blue{-0.06}) & 79.86 (\blue{-0.29}) & 4.76 & 39.0 \\
Net7 & convolution & self-attention & 76.24 (\blue{-1.07}) & 79.01 (\blue{-0.85}) & 4.83 & 45.0\\
\hline
ResNet-50 & \multicolumn{2}{c|}{network shape adjustment} & 77.43 (\red{+1.19}) & 78.73 (\blue{-0.28}) & 4.09 & 25.6\\
\hline
\end{tabular}
\end{center}
\label{tab:transition}
\end{table*}

\subsubsection{Replacing patch flattening with step-wise patch embedding}
\label{methodology:transition:flattening}

DeiT and ViT models directly encode the image pixels with a patch embedding layer which is equivalent to a convolution with a large kernel size and stride (\textit{e.g.}, 16). This operation flattens the image patches to a sequence of tokens so that Transformers can handle images. However, patch flattening impairs the position information within each patch and makes it more difficult to extract the patterns within patches. To solve this problem, existing methods usually attach a preprocessing module before patch embedding. The preprocessing module can be a feature extraction convnet~\cite{dosovitskiy2020image} or a specially designed Transformer~\cite{yuan2021tokens}.

We found that there is a rather simple solution, which is factorizing the large patch embedding to step-wise small patch embeddings. Specifically, We first add the stem layer in ResNet to the Transformer, which is a $7\times 7$ convolution layer with a stride of two. The stem layer can be seen as a $2\times 2$ patching embedding operation with pixel overlap (\textit{i.e.}, $7\times 7$ kernel size). Since the patch size in the original DeiT model is 16, we still need to embed $8\times 8$ patches after the stem. We further factorize the $8\times 8$ patch embedding to a $4\times 4$ embedding and a $2\times 2$ embedding, which are $4\times 4$ and $2\times 2$ convolution layers with stride 4 and 2 in the perspective of convolution. Additionally, we add an extra $2\times 2$ convolution to further upgrade the patch size from $16\times 16$ to $32\times 32$ before classification. These patch embedding layers can also be seen as the down-sampling layers and we double the channel numbers after embedding following the practice in convolution-based models.

By utilizing step-wise embeddings, the position prior within patches is encoded into features. As a result, the model can learn patterns more efficiently. As can be seen in Table~\ref{tab:transition}, this transition can significantly improve the base performance and elite performance of the network. It indicates that step-wise embedding is a better choice than larger patch embedding in Transformer-based models. Additionally, this transition is computationally efficient and only introduces about $4\%$ extra FLOPs.

\subsubsection{Stage-wise design}
\label{methodology:transition:stage}

In this section, we split networks into stages like ResNets. The blocks in the same stage share the same feature resolution. Since step-wise embeddings in the last transition have split the network into different stages, the transition in this section is to reassign the blocks to different stages as shown in Figure~\ref{fig:transition}. However, unlike convolution blocks, the complexity of self-attention blocks increases by $O\!\left(N^4\right)$ with respect to the feature size. Thus we only insert blocks to the $8\times8$, $16\times16$ and $32\times32$ patch embedding stages, which correspond to $28\times28$, $14\times14$ and $7\times7$ feature resolutions respectively for $224\times224$ inputs. Additionally, we halve the head dimension and feature dimension before self-attention in $28\times28$ stage to ensure that the blocks in different stages utilize similar FLOPs.

This transition leads to interesting results. The base performance is further improved. It is conjectured that the stage-wise design leverages the image local priors and thus can perform better under moderate augmentation. However, the elite performance of the network decreases markedly. To study reasons, we conduct ablation experiments and find that self-attention does not work well in very large resolutions. We conjecture that large resolution contains too many tokens and it is much more difficult for self-attention to learn relations among them. We will detail it in section~\ref{methodology:visformer}.

\subsubsection{Replacing LayerNorm with BatchNorm}
\label{methodology:transition:norm}

Transformer-based models usually normalize the features with LayerNorm~\cite{ba2016layer}, which is inherited from NLP tasks~\cite{vaswani2017attention, devlin2018bert}. As a contrast, convolution-based models like ResNets usually utilize BatchNorm~\cite{ioffe2015batch} to stabilize the training process.
LayerNorm is independent of batch size and more friendly for specific tasks compared with BatchNorm, while BatchNorm usually can achieve better performance given appropriate batch size~\cite{wu2018group}. 
We replace all the LayerNorm layers with BatchNorm layers and the results show that BatchNorm performs better than LayerNorm. It can improve both the base performance and elite performance of the network. 

In addition, we also try to add BatchNorm to Net2 to further improve the elite performance. However, this Net2-BN network suffers from convergence problems. This may explain why BatchNorm is not widely used in the pure self-attention models. But for our mixed model, BatchNorm is a reliable method to advance performance.

\subsubsection{Introducing $3\times3$ convolutions}
\label{methodology:transition:conv}

Since the tokens of the network are present as feature maps, it is natural to introduce convolutions with kernel sizes larger than $1\times1$. The specific meaning of large kernel convolution is illustrated at the bottom right of Figure~\ref{fig:transition}. When global self-attentions attempt to build the relations among all the tokens (\textit{i.e.}, pixels), convolutions focus on relating the tokens within local neighborhoods. We chose to insert $3\times 3$ convolutions between the $1\times1$ convolutions in feed-forward blocks, which transforms the MLP blocks into bottleneck blocks as exhibited at the top right of Figure~\ref{fig:transition}. Note that the channel numbers of the $3\times 3$ convolution layers are tuned to ensure that the FLOPs of the feed-forward blocks are nearly unchanged. The obtained bottleneck blocks are similar to the bottleneck blocks in ResNet-50, although they have different bottleneck ratios (\textit{i.e.}, the factor of reducing the channel numbers before the $3\times 3$ convolution). We replace the MLP blocks with bottleneck blocks in all three stages.

Not surprisingly, $3\times 3$ convolutions which can leverage the local priors in images further improve the network base performance. The base performance (77.37\%) becomes comparable with ResNet-50 (77.43\%). However, the elite performance decreases by 0.82\%. We conduct more experiments to study the reasons. Instead of adding $3\times 3$ convolutions to all stages, we insert $3\times 3$ convolutions to different stages separately. We observe that $3\times 3$ convolutions only work well on the high-resolution features. We conjecture that leveraging local relations is important for the high-resolution features in natural images. For the low-resolution features, however, local convolutions become unimportant when equipped with global self-attention. We will detail it in section~\ref{methodology:visformer}.

\subsubsection{Removing position embedding}
\label{methodology:transition:embedding}

In Transformer-based models, position embedding is proposed to encode the position information inter tokens. In the transition network, we utilize learnable position embedding as in~\cite{devlin2018bert} and add them to features after patch embeddings. To approaching ResNet-50, position embedding should be removed. 

The results are exhibited in Table~\ref{tab:transition}. The base performance is almost unchanged and the elite performance declines slightly (0.29\%). As a comparison, We test to remove the position embedding of DeiT-S and elite performance decreases significantly by 3.95\%. It reveals that position embedding is less important in the transition model than that in the pure Transformer-based models. It is because that the position prior inter tokens is preserved by the feature maps and convolutions with spatial kernels can encode and leverage it. Consequently, the harm of removing position embedding is remarkably reduced in the transition network. It also explains why convolution-based models do not need position embedding.

\subsubsection{Replacing self-attention with feed-forward}
\label{methodology:transition:feedforward}

In this section, we remove the self-attention blocks in each stage and utilize a feed-forward layer instead, so that the network becomes a pure convolution-based network. To keep the FLOPs unchanged, several bottleneck blocks are added to each stage. After the replacement, the obtained network consists of bottleneck blocks like ResNet-50. 

The performance of the obtained network (Net7) is shown in Table~\ref{tab:transition}. The pure convolution-based network performs much worse both in base performance and elite performance. \textit{It indicates that self-attentions do drive neural networks to higher elite performance and is not responsible for the poor base performance in ViT or DeiT. It is possible to design a self-attention network with high base performance and elite performance.} 

\subsubsection{Adjusting the shape of network}
\label{methodology:transition:shape}

There are still many differences between Net7 and ResNet-50. First, the shape of Net7 is different from ResNet-50. Their depths, widths, bottleneck ratios and block numbers in network stages are different. Second, they normalize the features in different positions. Net7 only normalizes input features in a block, while ResNet-50 normalizes features after each convolutional layer. Third, ResNet-50 down-samples the features with bottleneck blocks but Net7 utilizes a single convolution layer (\textit{i.e.}, patch embedding layer). In addition, Net7 employs a few more FLOPs. Nevertheless, both these two networks are convolution-based networks. The performance gap between these two networks can be attributed to architecture design strategy. 

As shown in Table~\ref{tab:transition}, the base performance is improved after transition. It demonstrates that ResNet-50 has better network architecture and can perform better with fewer FLOPs. However, ResNet-50 obtains worse elite performance. It indicates that the inconsistencies between base performance and elite performance exist not only in self-attention models but also in pure convolution-based networks. 

\subsection{Summary: the Visformer model}
\label{methodology:visformer}

We aim to build a network with high base performance and elite performance. The transition study has shown that there are some inconsistencies between base performance and elite performance. The first problem is the stage-wise design, which increases the base performance but decreases the elite performance. Stage-wise design re-arrange the blocks from one stage to three stages. Thus, for elite performance, some blocks in the new two stages must work less efficiently than those in the original stage. We replace the self-attention blocks with bottleneck blocks in each stage separately for Net5, by which we can estimate the importance of self-attention in different stages. The results are shown in Table~\ref{tab:self-attention-ablation}. The replacement of self-attention in all three stages reduces both the base performance and the elite performance. There is a trend that self-attentions in lower resolutions play more important roles than those in higher resolutions. Additionally, replacing the self-attentions in the first stage almost has no effect on the network performance. Larger resolutions contain much more tokens and we conjecture that it is more difficult for self-attentions to learn relations among them.

\begin{table}
\caption{Impact of replacing the self-attention blocks with the bottleneck blocks in each stage of Net5. These experiments are performed individually.}
\begin{center}
\begin{tabular}{|l|c|c|c|}
\hline
Network & base perf.(\%)& elite perf.(\%)\\
\hline\hline
Net5 & 77.37  & 80.15 \\
\hline
Net5-DS1  & 77.29 (\blue{-0.08}) & 80.13 (\blue{-0.02})\\
Net5-DS2 &77.34 (\blue{-0.02}) &79.75 (\blue{-0.40}) \\
Net5-DS3 &77.05 (\blue{-0.32}) &79.59 (\blue{-0.56})\\
\hline
\end{tabular}
\end{center}
\label{tab:self-attention-ablation}
\end{table}

\begin{table}
\begin{center}
\caption{Impact of replacing the MLP layers with the bottleneck blocks in each stage of Net4. These experiments are performed individually.}
\begin{tabular}{|l|c|c|c|}
\hline
Network & base perf.(\%)& elite perf.(\%)\\
\hline\hline
Net4 & 76.49  & 80.97 \\
\hline
Net4-S1  & 77.02 (\red{+0.53}) & 81.10 (\red{+0.13}) \\
Net4-S2  & 76.55 (\red{+0.06}) & 80.50 (\blue{-0.47})\\
Net4-S3  & 76.82 (\red{+0.33}) & 80.44 (\blue{-0.53})\\
\hline
Net5 & 77.37 (\red{+0.88}) & 80.15 (\blue{-0.82})\\
\hline
\end{tabular}
\end{center}

\label{tab:convolution-ablation}
\end{table}

\begin{table*}
\setlength{\tabcolsep}{0.08cm}
\caption{The configuration for constructing the Visformer-Ti and Visformer-S models, where `emb.' stands for feature embedding, and `s0'--`s3' indicate the four stages with different spatial resolutions.}
\newcommand{\tabincell}[2]{\begin{tabular}{@{}#1@{}}#2\end{tabular}}
\begin{center}
\begin{tabular}{|l|c|c|c|c|c|}
\hline
  & output size & Visformer-Ti & Visformer-S & VisformerV2-Ti & VisformerV2-S\\
\hline\hline

stem & $112\times112$ &  $7\times7$, $16$, stride $2$ &  $7\times7$, $32$, stride $2$ &  $7\times7$, $24$, stride $2$ &  $7\times7$, $32$, stride $2$ \\
\hline
emb. & $56\times56$ & - & - &  $2\times2$, $48$,  stride $2$ &  $2\times2$, $64$, stride $2$ \\
\hline
\multirow{5}{*}{s0} & \multirow{5}{*}{$56\times56$ } &\multirow{5}{*}{-} & \multirow{5}{*}{-} & \multirow{5}{*}{$\left[ \tabincell{c}{$1\times1$, $96$\\ $3\times3$, $96$\\ (group = $8$) \\ $1\times1$, $48$ \\} \right]$  $\times 1$  } &   \multirow{5}{*}{$\left[ \tabincell{c}{$1\times1$, $128$\\ $3\times3$, $128$\\(group = $8$) \\$1\times1$, $64$\\}\right]$  $\times 1$}\\
& & & & & \\
& & & & & \\
& & & & & \\
& & & & & \\
\hline
emb. & $28\times28$ &  $4\times4$, $96$,  stride $4$ &  $4\times4$, $192$, stride $4$ &  $2\times2$, $96$,  stride $2$ &  $2\times2$, $128$, stride $2$ \\
\hline
\multirow{5}{*}{s1} & \multirow{5}{*}{$28\times28$ }  &  
\multirow{5}{*}{$\left[ \tabincell{c}{$1\times1$, $192$\\ $3\times3$, $192$\\ (group = $8$) \\ $1\times1$, $96$ \\} \right]$  $\times 7$  } &  
\multirow{5}{*}{$\left[ \tabincell{c}{$1\times1$, $384$\\ $3\times3$, $384$\\(group = $8$) \\$1\times1$, $192$\\}\right]$  $\times 7$} & 
\multirow{5}{*}{$\left[ \tabincell{c}{$1\times1$, $192$\\ $3\times3$, $192$\\ (group = $8$) \\ $1\times1$, $96$ \\} \right]$  $\times 4$ }& 
\multirow{5}{*}{$\left[ \tabincell{c}{$1\times1$, $256$\\ $3\times3$, $256$\\(group = $8$) \\$1\times1$, $128$\\}\right]$  $\times 10$}\\
& & & & & \\
& & & & & \\
& & & & & \\
& & & & & \\
\hline
emb. & $14\times14$ &  $2\times2$, $192$,  stride $2$ &  $2\times2$, $384$, stride $2$ \\
\hline
\multirow{4}{*}{s2} & \multirow{4}{*}{$14\times14$ } &  
\multirow{4}{*}{$\left[ \tabincell{c}{MHSA, $192$\\ $1\times1$, $768$\\$1\times1$, $192$\\}\right]$ $\times 4$ }  &  
\multirow{4}{*}{$\left[ \tabincell{c}{MHSA, $384$\\ $1\times1$, $1536$\\$1\times1$, $384$\\}\right]$ $\times4$}  &
\multirow{4}{*}{$\left[ \tabincell{c}{MHSA, $192$\\ $1\times1$, $768$\\$1\times1$, $192$\\}\right]$ $\times 6$ }  &  
\multirow{4}{*}{$\left[ \tabincell{c}{MHSA, $256$\\ $1\times1$, $1024$\\$1\times1$, $256$\\}\right]$ $\times 14$}  \\
& & & & & \\
& & & & & \\
& & & & & \\
\hline
emb.& $7\times7$ &  $2\times2$, $384$,  stride $2$ &  $2\times2$, $768$, stride $2$ \\
\hline
\multirow{4}{*}{s3} & \multirow{4}{*}{$7\times7$ } &  
\multirow{4}{*}{$\left[ \tabincell{c}{MHSA, $384$\\ $1\times1$, $1536$\\ $1\times1$, $384$\\}\right]$ $\times4$ }  &    \multirow{4}{*}{$\left[ \tabincell{c}{MHSA, $768$\\ $1\times1$, $3072$\\$1\times1$, $768$\\}\right]$ $\times4$ } &
\multirow{4}{*}{$\left[ \tabincell{c}{MHSA, $384$\\ $1\times1$, $1536$\\ $1\times1$, $384$\\}\right]$ $\times2$ }  &    \multirow{4}{*}{$\left[ \tabincell{c}{MHSA, $512$\\ $1\times1$, $2048$\\$1\times1$, $512$\\}\right]$ $\times3$ } \\
& & & & & \\
& & & & & \\
& & & & & \\
\hline
& $1\times1$ & \multicolumn{4}{c|}{global average pool, 1000-d fc, softmax} \\
\hline
\multicolumn{2}{|c|}{FLOPs} & $1.3\times10^{9}$ & $4.9\times10^{9}$ & $1.3\times10^{9}$ & $4.3\times10^{9}$ \\
\hline
\end{tabular}
\end{center}
\label{tab:network-archtecture}
\end{table*}


The second problem is adding $3\times 3$ convolutions to the feed-forward blocks, which decreases the elite performance by 0.82\%. Based on Net4, we replace MLP blocks with bottleneck blocks in each stage separately. As can be seen in Table~\ref{tab:convolution-ablation}, although all stages obtain improvements in base performance, only the first stage benefits from bottleneck blocks in elite performance. The $3\times 3$ convolutions are not necessary for the other two low-resolution stages when self-attentions already have a global view in these positions. On the high-resolution stage, for which self-attentions have difficulty in handling all tokens, the $3\times 3$ convolutions can provide improvement.

Integrating the observation above, we propose the \textbf{Visformer} as vision-friendly, Transformer-based models. The detailed architectures are shown in Table~\ref{tab:network-archtecture}. Besides the positive transitions, Visformer adopts the stage-wise design for higher base performance. But self-attentions are only utilized in the last two stages, considered that self-attention in the high-resolution stage is relatively inefficient. Visformer employs bottleneck blocks in the first stage and utilizes group $3\times 3$ convolutions in bottleneck blocks inspired by ResNeXt~\cite{xie2016aggregated}. We also introduce BatchNorm to patch embedding modules as in CNNs. We name Visformer-S to denote the model that directly comes from DeiT-S. In addition, we can adjust the complexity by changing the output dimensionality of multi-head attentions. Here, we shrink the dimensionality by half and derive the Visformer-Ti model, which requires around $1/4$ computational costs of the Visformer-S model.

\subsection{VisformerV2: optimizing the architecture configuration}
Some architecture configurations of Visformer are not carefully tuned. For example, when splitting the network into different stages, we averagely assign the 12 blocks to the three stages, expect that we utilize 3 more blocks in the first stage to compensate for the removal of self-attention. In other words, the stage configuration ([7, 4, 4]) is not carefully designed. Furthermore, depth and width, which are also not polished in Visformer, have been demonstrated to be very important configurations for network performance. Therefore, we conduct many experiments to explore the architecture of Visformer and propose VisformerV2. VisformerV2 is much better than the original Visformer and the architecture is shown in Table~\ref{tab:network-archtecture}. The detailed analysis and experiments are shown in Section~\ref{deisgning_v2}.

\subsection{Transformer with Half-precision}
Recently, quantization has been widely used to accelerate the training process and save GPU memory. Specifically, half-precision floating-point (FP16), the lowest precision that can preserve the network performance, has been adopted by many researchers. However, some works~\cite{touvron2020training, ding2021cogview, liu2021swinv2} have shown that half-precision can lead to overflows in Transformers and we also observe this problem in Visformer. 

\begin{table}[b!]
\caption{Comparison of inference time for VisformerV2-S with different score generating methods. The tested GPU is V100 and the batch size is 32.}
\begin{center}
\begin{tabular}{|l|c|c|c|}
\hline
Method & original & PB-Relax & ours \\
\hline
Batch Time (ms) & 42.8  & 46.6 & 43.0\\

\hline
\end{tabular}
\end{center}
\label{tab:different-sa}
\end{table}

 Based on our experimental analysis, we find that attention score generation can cause overflow. With the queries ($Q$) and keys ($K$), the standard self-attention scores can be computed as:
\begin{gather}
    A_{score} = \mathrm{softmax}(\frac{QK^{T}}{\sqrt{d}})
\end{gather}
However, $Q$ and $K$ can be very large matrices and the elements in $QK^{T}$ will be the dot-product of two very long vectors. As a result, the scores can overflow easily while utilizing 16-bit precision. To solve this problem, we first try to pre-normalize $Q$ and $K$:
\begin{gather}
    A_{score} = \mathrm{softmax}( (\frac{Q}{\sqrt[4]{d}}) (\frac{K^{T}}{\sqrt[4]{d}} ))
\end{gather}
Where $d$ is the length of the vector. Nevertheless, the attention scores still overflow sometimes. This is because the scores are only normalized with $\sqrt{d}$ overall. As the dot product of two vectors with length $d$, the scores are still under the risk of overflow. Consequently, we try to normalize the score with $d$:
\begin{gather}
    A_{score} = \mathrm{softmax}( (\frac{Q}{\sqrt{d}}) (\frac{K^{T}}{\sqrt{d}} ))
\end{gather}
In our experiments, we observed that it can effectively avoid overflow during computing scores and will not degrade the network performance.

Note that CogView~\cite{ding2021cogview} also proposes PB-Relax to eliminate overflow in attention scores. PB-Relax pre-minuses the maximum of the attention scores. However, this method needs to tune a hyper-parameter and usually considerably increases the network runtime, as shown in Table~\ref{tab:different-sa}. As a contrast, our method nearly does not introduce extra runtime.

\section{More Experiments on Visformer}

\subsection{The improvements on the upper-bound and lower-bound}
We first compare Visformer against DeiT, the direct baseline. Results are summarized in Table~\ref{tab:visformer-model}. Using comparable computational costs, the Visformer models outperform the corresponding DeiT models significantly. Specifically, the advantages of Visformer-S and Visformer-Ti over DeiT-S and DeiT-Ti under the elite setting are 2.12\% and 6.41\%, while under the base setting, the numbers grow to 14.08\% and 10.47\%, respectively. In other words, the advantage becomes more significant under the base setting, which is more frequently used for visual recognition.

\begin{table}
\caption{The comparison of base and elite performance as well as the FLOPs between Visformer and DeiT, the direct baseline.}
\newcommand{\tabincell}[2]{\begin{tabular}{@{}#1@{}}#2\end{tabular}}
\begin{center}
\begin{tabular}{|l|c|c|c|}
\hline
Network & \tabincell{c}{base perf.\\(\%)} & \tabincell{c}{elite perf.\\(\%)} & \tabincell{c}{FLOPs\\(G)}\\
\hline\hline
Visformer-Ti &74.34 & 78.62  & 1.3\\
DeiT-Ti & 63.87 & 72.21 & 1.3\\
\hline
Visformer-S & 77.20 & 82.19  & 4.9  \\
DeiT-S  & 63.12 & 80.07  & 4.6  \\
\hline
\end{tabular}
\end{center}
\label{tab:visformer-model}
\end{table}

\subsection{Training with limited data}
We evaluate the performance of Visformer in the scenario with limited training data, which we consider is an important ability of being vision-friendly, while prior Transformer-based models mostly required abundant training data~\cite{dosovitskiy2020image}.

Four subsets of ImageNet are used, with 10\% and 1\% randomly chosen classes (all data), and with 10\% and 1\% randomly chosen images (all classes), respectively. To challenge the models, we still use the elite setting with 300 epochs (not extended). As shown in Table~\ref{tab:limited_data}, it is observed that the DeiT-S model reports dramatic accuracy drops in all the four tests (note that the accuracy of using only 10\% and 1\% classes should be much higher if epochs are extended). In comparison, Visformer remains robust in these scenarios, showing its potential of being used for visual recognition with limited data.

In tiny level, ResNet-50-55\% is obtained by reducing the channel numbers (like other tiny models) to 55\% (so that the FLOPs, 1.3G, is similar to Visformer-Ti and Deit-Ti). The conclusion is similar: Visformer-Ti is still the best overall model, and the advantage is slightly enlarged because the risk of over-fitting has been reduced.

\begin{table}
\caption{Comparison among Visformer, DeiT, and ResNet, in terms of classification accuracy (\%) using limited training data. The elite setting with 300 epochs is used for all models.}
\setlength{\tabcolsep}{0.12cm}
\newcommand{\tabincell}[2]{\begin{tabular}{@{}#1@{}}#2\end{tabular}}
\begin{center}
\begin{tabular}{|l|c|c|c|c|c|}
\hline
Network & \tabincell{c}{100\%\\classes} & \tabincell{c}{10\%\\classes} & \tabincell{c}{1\%\\classes} &\tabincell{c}{10\%\\images}&\tabincell{c}{1\%\\images}\\
\hline\hline
DeiT-S & 80.07 & 80.06 & 73.40 & 40.41 &6.94\\
ResNet-50 & 78.73 & 89.90 &93.20 &58.37 &13.59 \\
Visformer-S &82.19 & 90.06  & 91.60 &58.74 &16.56\\
\hline
Deit-Ti & 72.33 & 78.72 & 74.40 & 38.44 &6.53\\
ResNet-50-55\% & 72.84 & 87.10 &91.40 &51.48 &10.68 \\
Visformer-Ti & 78.62 & 89.48  & 90.60 &55.14 &11.79\\
\hline
\end{tabular}
\end{center}
\label{tab:limited_data}
\end{table}

\subsection{Designing VisformerV2}
\label{deisgning_v2}

\begin{table*}[t]
\caption{The elite performance and inference time of different Visformer models. The batch time is tested on a V100 GPU with a batch size of 32.}
\newcommand{\tabincell}[2]{\begin{tabular}{@{}#1@{}}#2\end{tabular}}
\begin{center}
\begin{tabular}{|l|l|c|c|c|c|c|}
\hline
Network & block numbers & channel numbers &  \tabincell{c}{elite perf.(\%)} & \tabincell{c}{FLOPs (G)} & \tabincell{c}{Params (M)} & \tabincell{c}{Batch Time (ms)}\\
\hline\hline
Visformer-S & \{0, 7, 4, 4\} & \{96, 192, 384, 768\}  & 82.39 & 4.9 & 40.2 & 36.9\\
\hline
 & \{1, 6, 4, 4\} &  \multirow{2}{*}{\{96, 192, 384, 768\}}  & 82.37 & 4.9 & 40.3 & 37.3\\
\cline{2-2}\cline{4-7}
  & \{3, 4, 4, 4\} &    & 81.70 & 4.9 & 39.3 & 38.4\\
\cline{2-7}
 & \{1, 3, 11, 11\} &  \multirow{4}{*}{\{64, 128, 256, 512\}} & 81.73 & 4.2 & 45.4 & 42.1\\
 \cline{2-2}\cline{4-7}
 & \{1, 6, 10, 10\} &    & 82.20 & 4.2 & 41.9 & 41.7\\
 \cline{2-2}\cline{4-7}
 & \{1, 3, 18, 3\} &     & 82.51 & 4.2 & 25.8 & 43.4\\
 \cline{2-2}\cline{4-7}
 & \{1, 6, 16, 3\} &    & 82.89 & 4.2 & 24.6 & 42.8\\
\hline
 VisformerV2-S& \{1, 10, 14, 3\} & \{64, 128, 256, 512\}  & 82.97 & 4.3 & 23.6 & 43.0\\
 \hline
\end{tabular}
\end{center}
\label{tab:designing_v2}
\end{table*}

We design VisformerV2 by polishing the original Visformer. First, we apply relative position bias~\cite{liu2021swin} to Visformer, which improves the results to 82.39\% as shown in Table~\ref{deisgning_v2}. Then we test whether we need to utilize an extra early stage. As shown in Table~\ref{tab:designing_v2}, assigning a block to the new stage does not improve the performance and furthermore, the performance decreases when more blocks are assigned to it. However, we find that this stage can improve the detection and segmentation results, which will be detailed in Section~\ref{down-stream-tasks}. Therefore, we decide to utilize one block in the new stage. Next, we test to utilize deep and narrow architecture~\cite{dong2021cswin}. We first narrow down the network and directly assign blocks to the self-attention stages (the last two stages). The tested stage configurations are \{1, 3, 11, 11\} and \{1, 6, 10, 10\}. These settings degrade the performance. Then we try to assign more blocks to the third stage ( \textit{i.e.}, \{1, 3, 18, 3\} and \{1, 6, 16, 3\}), which is the default setting for many convolution~\cite{he2016deep} and Transformer networks~\cite{liu2021swin, dong2021cswin}. It improves the networks significantly. We also find that the second pure convolution stage is very important. Moving the blocks from this stage to the other stages will substantially degrade the network. Therefore we assign more blocks to this stage and obtain VisformerV2-S. With a similar study, we design VisformerV2-Ti. The detailed architecture is shown in Table~\ref{tab:network-archtecture}.

Note that the deep and narrow architecture significantly increases the runtime on GPU. This is because that the wide and shallow architecture has a better parallelization property. To compensate for the loss in runtime, we utilize fewer FLOPs for `deep-narrow' networks. More importantly, we find that when the input resolution is enlarged (detection and segmentation tasks in Section~\ref{down-stream-tasks}) or the model is scaled up, the parallelization property will be improved and the runtime on GPU becomes more consistent with FLOPs. 

\begin{table}
\caption{Comparison among our method and other Transformer-based vision models. `*' indicates that we re-run the model using the elite setting. `KD' stands for knowledge distillation~\cite{hinton2015distilling}.}
\newcommand{\tabincell}[2]{\begin{tabular}{@{}#1@{}}#2\end{tabular}}
\begin{center}
\begin{tabular}{|l|c|c|c|}
\hline
Methods & Top-1(\%) & \tabincell{c}{FLOPs\\(G)} & \tabincell{c}{Params\\(M)} \\
\hline\hline
ResNet-18~\cite{he2016deep} & 69.8 & 1.8  & 11.7  \\
DeiT-Ti~\cite{touvron2020training} & 72.2 & 1.3  & 5.7  \\
DeiT-Ti (KD)~\cite{touvron2020training} & 74.6 & 1.3  & 5.7  \\
AutoFormer-Ti~\cite{chen2021autoformer} & 74.7 & 1.3 & 5.7 \\
PVT-Ti~\cite{wang2021pyramid} & 75.1  & 1.9 & 13.2  \\
PVTv2-B1~\cite{wang2021pvtv2} & 78.7 & 2.1 & 13.1 \\
\textbf{Visformer-Ti (ours)} & 78.6 & 1.3 & 10.3 \\
\textbf{VisformerV2-Ti (ours)} & 79.6 & 1.3 & 9.4 \\
\hline\hline
ResNet-50~\cite{he2016deep} & 76.2 & 4.1  & 25.6  \\
ResNet-50$^*$~\cite{he2016deep} & 78.7 & 4.1  & 25.6  \\
RegNetY-4GF~\cite{radosavovic2020designing} & 79.4 & 4.0  & 20.6  \\
RegNetY-8GF~\cite{radosavovic2020designing} & 79.9 & 8.0  & 39.2  \\
RegNetY-4GF$^*$~\cite{radosavovic2020designing} & 80.0 & 4.0  & 20.6  \\
\hline
DeiT-S~\cite{touvron2020training} &79.8 & 4.6 & 21.8    \\
DeiT-S$^*$~\cite{touvron2020training} & 80.1 & 4.6 & 21.8   \\
DeiT-B~\cite{touvron2020training} & 81.8 & 17.4 &86.3 \\
\hline
PVT-S~\cite{wang2021pyramid} & 79.8 & 3.8 & 24.5 \\
PVT-Medium~\cite{wang2021pyramid} & 81.2 & 6.7 & 44.2 \\
PVTv2-B2-Li~\cite{wang2021pvtv2} & 82.1 & 3.9 & 22.6 \\
PVTv2-B2~\cite{wang2021pvtv2} & 82.0 & 4.0 & 25.4 \\
\hline
Swin-T~\cite{liu2021swin} & 81.3 & 4.5 & 29 \\
\hline
CvT-13~\cite{wu2021cvt} & 81.6 & 4.5 & 20 \\
CvT-13-NAS~\cite{wu2021cvt} & 82.2 & 4.1 & 18\\
CvT-13(384)~\cite{wu2021cvt} & 83.0 & 16.3 & 20\\
\hline
Conformer-Ti~\cite{peng2021conformer} & 81.3 & 5.2 & 23.5 \\
\hline
T2T-ViT$_t$-14~\cite{yuan2021tokens} & 80.7& 5.2 & 21.5\\
T2T-ViT$_t$-19~\cite{yuan2021tokens}& 81.4 & 8.4 & 39.0\\
\hline
BoTNet-S1-59~\cite{srinivas2021bottleneck} & 81.7 & 7.3 & 33.5 \\
\hline
CSWin-T~\cite{dong2021cswin} & 82.7 & 4.3 & 23 \\
\hline
AutoFormer-S~\cite{chen2021autoformer} & 81.7 & 5.1 & 22.9 \\
\hline
\textbf{Visformer-S (ours)} & 82.2 & 4.9 & 40.2 \\
\textbf{VisformerV2-S (ours)} & 83.0 & 4.3 & 23.6 \\
\hline
\end{tabular}
\end{center}
\label{tab:comparison-with-sota} 
\end{table}

\begin{table}
\caption{Comparison of inference efficiency among Visformer and other models on a 32G-V100. A batch size of $32$ is used for testing. `*' indicates that the model is re-trained with the elite setting.}
\newcommand{\tabincell}[2]{\begin{tabular}{@{}#1@{}}#2\end{tabular}}
\begin{center}
\begin{tabular}{|l|c|c|c|}
\hline
Methods & \tabincell{c}{Top-1\\(\%)} & \tabincell{c}{FLOPs\\(G)} & \tabincell{c}{Batch Time\\(ms)} \\
\hline\hline
ResNet-50$^*$ & 78.7 & 4.1  & 34.2   \\
DeiT-S$^*$ & 80.1 & 4.6 & 36.9   \\
RegNetY-4GF$^*$ & 80.0 & 4.0  & 40.2   \\
Swin-T & 81.3 & 4.5 & 47.6   \\
CSwin-T & 82.7 & 4.3 & 57.5   \\
PVT-S & 79.8 & 3.8 & 47.6 \\
PVTv2-B2 & 82.0 & 4.0 & 57.1 \\
PVTv2-B2-Li & 82.1 & 3.9 & 56.8 \\
\hline
EfficientNet-B3~\cite{tan2019efficientnet} & 81.6 & 1.8 & 48.3 \\
EfficientNet-B4~\cite{tan2019efficientnet} & 82.9 & 4.2 & 81.7 \\
\hline
Visformer-S (ours)& 82.2 & 4.9 & 36.7 \\
VisformerV2-S (ours)& 83.0 & 4.3 & 43.0 \\
\hline

\end{tabular}
\end{center}
\label{tab:computing-time}
\end{table}

\subsection{Comparison to the state-of-the-arts}
We then compare Visformer and VisformerV2 to other Transformer-based approaches in Table~\ref{tab:comparison-with-sota}. At the tiny level, Visformer-Ti and VisformerV2-Ti outperform other vision Transformers that with similar FLOPs. For larger models, Visformer-S performs much better than most of the models with similar FLOPs. VisformerV2-S further improves the performance and outperforms other vision Transformer models. Note that VisformerV2-S utilize fewer FLOPs and parameters than Visformer-S.

\subsection{Inference efficiency}
Although VisformerV2-S is not as efficient as Visformer-S in runtime, it is still much faster than most vision Transformer models as shown in Table~\ref{tab:computing-time}. As for the state-of-the-art EfficientNet convnets, Visformer-S are below the EfficientNets with similar FLOPs. However, EfficientNets are computing inefficient on GPUs. It is shown that Visformer-S is significantly faster than EfficientNet-B3 which performance is slightly worse than our model. VisformerV2-S and EfficientNet-B4 have similar FLOPs and performance, but VisformerV2-S is significantly faster than EfficientNet-B4.


\subsection{COCO Object Detection}
\label{down-stream-tasks}
Last but not least, we evaluate our models on the COCO object detection task. Since the standard self-attention in Visformer models is not efficient for high-resolution inputs, we simply replace self-attention with the shifted window (Swin) self-attention~\cite{liu2021swin} to apply our models to the detection task. Therefore, Swin Transformers are our important baseline models. It should be emphasized that the self-attention in Visformer can also be replaced with other resolution-friendly self-attentions like CSWin self-attention and MSG self-attention. We just utilize the widely used Swin self-attention to show the superiority of Visformer architecture. 

The models are evaluated with two frameworks: Mask-RCNN~\cite{he2017mask} and Cascade Mask-RCNN~\cite{cai2018cascade}. We train the models on COCO 2017 dataset and report the results on COCO val2017. We inherit the training settings in~\cite{liu2021swin}: the AdamW optimizer with a learning rate of 0.0001 and the weight decay of 0.05. The batch size is 16 and we show the results of $1\times$ (12 epochs) and $3\times$ (36 epochs) schedule. The FPS is measured on a V100 GPU with a batch size of 1. The FLOPs are computed with $1280\times800$ resolution.

We first test different methods with the Mask R-CNN $1\times$ schedule. As shown at the top of Table~\ref{tab:detection}, Visformer-S slightly outperform Swin-T. When we assign a block to the first stage (Visformer-S-F), although the classification performance is not improved (illustrated in Section~\ref{deisgning_v2}), the detection result becomes better. We conjecture that the block in the first stage can help the FPN~\cite{lin2017feature} to explore the low-level features. The VisformerV2-S further improves the performance and outperforms Swin-T by 2.2\%. Additionally, because of the improvement on parallelization property, the FPS becomes consistent with FLOPs and VisformerV2-S is faster than Visformer-S. 

For the Cascade Mask R-CNN framework, VisformerV2-S still outperforms Swin-T by a large margin. We compare VisformerV2-S with more methods for $3\times$ and multi-scale schedule~\cite{carion2020end,sun2021sparse}, and our model still performs better than the other methods. As for FPS, our method is as efficient as Swin-T and MSG-T, and is faster than other vision Transformer Methods. 

\begin{table*}
\setlength{\tabcolsep}{0.12cm}
\newcommand{\tabincell}[2]{\begin{tabular}{@{}#1@{}}#2\end{tabular}}
\begin{center}
\begin{tabular}{|l|c|c|c|c|c|c|c|c|c|c|}
\hline
Method & \tabincell{c}{Backbone} & $\mathrm{AP^{box}}$ &  $\mathrm{AP^{box}_{50}}$ & $\mathrm{AP^{box}_{75}}$ & $\mathrm{AP^{mask}}$ &  $\mathrm{AP^{mask}_{50}}$ & $\mathrm{AP^{mask}_{75}}$ & FLOPs & Params & FPS \\
\hline\hline
\multirow{5}{*}{\tabincell{l}{Mask R-CNN \\ $1\times$ schedule} }& R-50 & 38.0 & 58.6  & 41.4 & 34.4 & 55.1 & 36.7 & 260 & 44 & 18.6\\
  & Swin-T &42.6 & 65.1  & 46.2 & 39.3 & 62.0 & 42.1 & 267 & 48 &14.8 \\
  & Visformer-S &43.0 & 65.3  & 47.2 &39.6 &62.4 & 42.4 & 275 & 60 &13.1 \\
  & Visformer-S-F & 43.5 & 65.9 & 47.7 & 39.8 & 62.5 & 42.6 & 275 & 60 & 13.0 \\
  & VisformerV2-S &44.8 & 66.8  & 49.4 & 40.7 & 63.9 & 43.7 & 262 & 43 & 15.2\\
\hline 
\multirow{3}{*}{\tabincell{l}{Mask R-CNN \\ $3\times$ + MS schedule} } & R-50 & 41.0 & 61.7  & 44.9 &37.1 & 58.4 & 40.1 & 260 & 44 & 18.6\\
  & Swin-T &46.0 & 68.2  & 50.2 & 41.6 & 65.1 & 44.8 & 267 & 48 &14.8\\
  & VisformerV2-S &47.8 & 69.5  & 52.6 & 42.5 & 66.4 & 45.8& 262 & 43 & 15.2\\
\hline 
  \multirow{3}{*}{\tabincell{l}{Cascade Mask R-CNN \\ $1\times$ + MS schedule} }& R-50 &43.7 & 61.7  & 47.5 &38.0 &58.8 & 41.0 &739 & 82&10.6\\
  & Swin-T &48.1 & 67.1  & 52.2 & 41.7 & 64.4 & 45.0 & 745 & 86 & 9.5\\
  & VisformerV2-S &49.3 & 68.1  & 53.6 & 42.3 & 65.1 & 45.7 & 740 & 81 &9.6\\
\hline 
 \multirow{7}{*}{\tabincell{l}{Cascade Mask R-CNN  \\ $3\times$ + MS schedule} }  & R-50 & 46.3 & 64.3  & 50.5 & 40.1 & 61.7 & 43.4 & 739 & 82&10.6\\
  &DeiT-S & 48.0 & 67.2 & 51.7 & 41.4 & 64.2 & 44.3 & 889 & 80 & - \\
  & Swin-T &50.5 & 69.3  & 54.9 & 43.7 & 66.6 & 47.1& 745 & 86 & 9.5\\
  &MSG-T~\cite{fang2021msg} & 50.3 &69.0 & 54.7 & 43.6 & 66.5 & 47.5 & 758 & 86 & 9.5 \\
  &PVTv2-B2-Li~\cite{wang2021pvtv2} &50.9 & 69.5 & 55.2 & - & - & - & 725 & 80 & 8.2 \\
  &PVTv2-B2~\cite{wang2021pvtv2} & 51.1 & 69.8 & 55.3 & - & - & - &  788 & 83& 7.1 \\
  & VisformerV2-S &51.6 & 70.1 & 56.4 & 44.1 & 67.5 & 47.8 & 740 & 81 &9.6\\
\hline
\end{tabular}
\end{center}
\caption{Object detection and instance segmentation performance on COCO 2017. The FPS is measured on a V100 GPU with a batch size of 1. The FLOPs are computed with $1280\times800$ resolution. `MS' indicates multi-scale training~\cite{carion2020end,sun2021sparse}}.
\label{tab:detection}
\end{table*}

\section{Conclusions}

This paper presents Visformer, a Transformer-based model that is friendly to visual recognition. We propose to use two protocols, the base and elite setting, to evaluate the performance of each model. To study the reason why Transformer-based models and convolution-based models behave differently, we decompose the gap between these models and design an eight-step transition procedure that bridges the gap between DeiT-S and ResNet-50. By absorbing the advantages and discarding the disadvantages, we obtain the Visformer-S model that outperforms both DeiT-S and ResNet-50. Visformer also shows a promising ability when it is transferred to a compact model and when it is evaluated on small datasets.


\section{Acknowledgments}
\noindent This work was supported by the National Key R\&D Program of China (2017YFB1301100), National Natural Science Foundation of China (61772060, U1536107, 61472024, 61572060, 61976012, 61602024), and the CERNET Innovation Project (NGII20160316).

{
\bibliographystyle{IEEEtran}
\bibliography{visformer_jrnl}
}

\vfill

\end{document}